\documentclass[conference]{IEEEtran}
\IEEEoverridecommandlockouts
\usepackage{amsmath,amssymb,amsfonts}
\usepackage{algorithmic}
\usepackage{graphicx}
\usepackage{textcomp}
\usepackage{xcolor}
\usepackage{subfigure}
\usepackage{subcaption}
\def\BibTeX{{\rm B\kern-.05em{\sc i\kern-.025em b}\kern-.08em
    T\kern-.1667em\lower.7ex\hbox{E}\kern-.125emX}}

\usepackage{tikz}
\def\checkmark{\tikz\fill[scale=0.4](0,.35) -- (.25,0) -- (1,.7) -- (.25,.15) -- cycle;}

\begin{document}

\title{Towards Learning Scalable Agile Dynamic Motion Planning for Robosoccer Teams with Policy Optimization\\
\thanks{This work was supported by the Office of Naval Research under grant N00014-22-1-2834 
}
}

\author{Brandon Ho*,
Batuhan Altundas, 

Matthew Gombolay

}

\maketitle

\begin{abstract}
In fast-paced, ever-changing environments, dynamic Motion Planning for Multi-Agent Systems in the presence of obstacles is a universal and unsolved problem. Be it from path planning around obstacles to the movement of robotic arms, or in planning navigation of robot teams in settings such as Robosoccer, dynamic motion planning is needed to avoid collisions while reaching the targeted destination when multiple agents occupy the same area. In continuous domains where the world changes quickly, existing classical Motion Planning algorithms such as RRT* and A* become computationally expensive to rerun at every time step. Many variations of classical and well-formulated non-learning path-planning methods have been proposed to solve this universal problem but fall short due to their limitations of speed, smoothness, optimally, etc. Deep Learning models overcome their challenges due to their ability to adapt to varying environments based on past experience. However, current learning motion planning models use discretized environments, do not account for heterogeneous agents or replanning, and build up to improve the classical motion planners' efficiency, leading to issues with scalability.  To prevent collisions between heterogenous team members and collision to obstacles while trying to reach the target location, we present a learning-based dynamic navigation model and show our model working on a simple environment in the concept of a simple Robosoccer Game. 
\end{abstract}
\vspace{-0.3cm}
\section{Introduction}
A fast dynamic Motion Planning for Multi-Agent Systems in the presence of obstacles is a universal and unsolved problem \cite{gonzalez_review_2016, grigorescu_survey_2020}  applicable to robotics, planning, and optimization. 
In many industrial applications, users are interested in trajectories with minimum time to achieve the highest productivity \cite{paden_survey_2016}. Especially in athletics, Agile Motion Planning becomes an important factor for fast reactions for both gameplay quality and safety \cite{zaidi_athletic_2023}, especially when collaborating with humans. Our work can be used to provide an agile motion planning policy for Robosoccer, an environment that aims to promote AI and robotic research, specifically in a soccer domain  \cite{kitano_robocup_1998}.

In continuous domains where the world changes quickly, existing classical Motion Planning algorithms such as RRT*  \cite{karaman_sampling-based_2011} and A* \cite{duchon_path_2014} become computationally expensive to rerun at every time step. 
The classical algorithms cannot capture the real-world dynamics to produce a non-collision path at once for all future time steps. Many variations of classical and well-formulated non-learning path-planning methods have been proposed to solve this universal problem \cite{karaman_sampling-based_2011} \cite{lavalle_rapidly-exploring_nodate} \cite{karaman_incremental_2010}. Those methods are generalized in the following categories: sampling-based, graph-based, geometric-based, and optimization-based. However, these methods have their disadvantages. Sampling-based methods lack in producing consistent trajectories, and the produced trajectories are not smooth \cite{gonzalez_review_2016, grigorescu_survey_2020}. 

Deep learning models’ ability to adapt based on past experiences to varying environments is more appealing to use than classical models with fixed parameters \cite{teng_motion_2023}. Currently, the application of deep learning in motion planning is unmatured and will continue to grow, especially in Multi-Agent Reinforcement Learning (MARL) \cite{grigorescu_survey_2020, huang_autonomous_2020, teng_motion_2023}. Many current learning motion planners have similar characteristics: discretized environment, single- or multi-agent, non-generative, and no replanning. A discretized environment limits the learning algorithm's efficiency and cost-effectiveness based on the granularity of the discretization. Many learning models do not account for heterogeneous agents.  Learning-based approaches such as Neural-RRT \cite{wang_neural_2020} or Neural-A* \cite{yonetani_path_2021} utilize a Deep Learning method to provide candidate locations or probability distributions for classical algorithms to leverage instead of generating a path themselves. Thus, they still run the baseline algorithms to generate the paths, making them inefficient. These methods, while shown to be faster than simple RRT* and A*, respectively, are still slow compared to pure learning-based approaches.


Using a Learning-based Motion Planner with symbolic knowledge of the agents provides a new path-finding method that can be trained to maximize the number of targets reached while avoiding collisions. In addition, our method minimizes the computational cost of recalculating a new non-colliding path in a dynamic environment.


We present the following in this paper:
\begin{itemize}
    \item Present a Dynamic Motion Planning Environment for Heterogenous Teams that can be used for Robots in Sports.
    \item Provide an End-to-End Trainable Method that can be run in a Decentralized Setting for Agile-Motion Planning.
    \item Discuss future work to make the models scalable through the use of Graph Neural Networks to allow for the use of the model in different team compositions.
\end{itemize}

\newpage

\section{Background}

\begin{table}[ht] 
\resizebox{0.9\columnwidth}{!}{
\begin{minipage}{\columnwidth}
\begin{tabular}{|c|c|c|c|c|c|c|c|}
\hline
 & \textbf{\begin{tabular}[c]{@{}c@{}}\rotatebox[origin=c]{90}{ Discrete/Continuous } \rotatebox[origin=c]{90}{Space}\end{tabular}} & \textbf{\rotatebox[origin=c]{90}{Multi-Agents}} & \textbf{\begin{tabular}[c]{@{}c@{}} \rotatebox[origin=c]{90}{Scalability}\end{tabular}} & \textbf{\begin{tabular}[c]{@{}c@{}}\rotatebox[origin=c]{90}{Generative} \rotatebox[origin=c]{90}{Motion Planner}\end{tabular}} & \textbf{\rotatebox[origin=c]{90}{Replanning}} & \textbf{\begin{tabular}[c]{@{}c@{}}\rotatebox[origin=c]{90}{Dynamic Obstacles} \end{tabular}} & \textbf{\begin{tabular}[c]{@{}c@{}}\rotatebox[origin=c]{90}{Heterogeneous} \rotatebox[origin=c]{90}{Agents}\end{tabular}} \\ \hline
Keselman et al. 2018\cite{keselman_reinforcement_2018}& \textbf{D} & \textbf{} & \textbf{} &&&&\\ \hline
Wang et al. 2020\cite{wang_neural_2020} & \textbf{D} & \textbf{} & \textbf{} &&&&\\\hline
Yonetani et al. 2021\cite{yonetani_path_2021}& \textbf{D} & \textbf{} & \textbf{} &&&&\\ \hline
Lv et al. 2019\cite{lv_path_2019} & \textbf{D} & \textbf{} & \textbf{} & \textbf{\checkmark} &&&  \\ \hline
Ichter et al. 2018\cite{ichter_learning_2018} & \textbf{C} & \textbf{} & \textbf{} & \textbf{} &&&  \\ \hline
Khan et al. 2020\cite{khan_graph_2020}& \textbf{C} &  & \textbf{\checkmark} &&&&  \\ \hline
Yu, Gao, et al, 2021\cite{yu_reducing_2021}& \textbf{C} & \textbf{} & \textbf{\checkmark} & \textbf{\checkmark} &&&  \\ \hline
Li et al. 2020\cite{li_graph_2020} & \textbf{D} & \textbf{\checkmark} & \textbf{} & \textbf{\checkmark} & \textbf{\checkmark} &&  \\ \hline
Liu et al. 2020\cite{liu_mapper_2020} & \textbf{D} & \textbf{\checkmark} &&& \textbf{\checkmark} & \textbf{\checkmark} &  \\ \hline
Zhang et al. 2022\cite{zhang_learning-based_2022}& \textbf{C} && \textbf{\checkmark} & \textbf{\checkmark} & \textbf{} & \textbf{\checkmark} &  \\ \hline
\textbf{Ours} & \textbf{C} & \textbf{\checkmark} & \textbf{ } & \textbf{\checkmark} & \textbf{\checkmark} & \textbf{\checkmark} & \textbf{\checkmark} \\ \hline
\end{tabular}
\caption{Related Works in Motion Planning and their applications of Domain, Scalability, Methods and Team Compositions.}
\label{tab:related_works}
\end{minipage}
}
\end{table}



\vspace{-0.5cm}

Table \ref{tab:related_works} represents recent work on motion planning. The existing work indicate that there is a need for heterogenous multi-agent motion planner for dynamic environments in both discrete and continuous space. While this paper focuses on the learning of a dynamic motion planner, we plan to address scalability, which accounts for either multiple obstacles or multiple agents, in our future work.

\subsection{Motion Planning}
The simplest motion planning goes by in a straight line from a start location to the end location with no consideration of obstacles. Non-learning motion planning algorithms, including RRT, RRT*, and RRG, account for obstacle avoidance but have poor sample efficiency. Learning motion planning algorithms, such as a learning RRT, can improve the sample efficiency of its non-learning counterpart. However, learning motion planner algorithms still require non-learning motion planners to generate their path, so they have the same time complexity as their non-learning counterparts.

\color{black}



\section{Problem Statement and Setup}
\subsection{Environment}

We present a simple motion planning environment that allows for the representation of team members, opponents, and goal locations. The environment can be translated into the context of Robosoccer Environment\cite{kitano_robocup_1998} as shown in Fig. \ref{fig:sample_plan}. The agent is tasked with moving to the target location with the ball, intercepting it, and moving it to another series of locations while avoiding getting close to the opponent players who aim to intercept the player or avoid colliding with the other teammates. Both other teammates and opponent players are represented as static or moving obstacles, making the model decentralized. The training was done on a single agent, and therefore each agent can also be trained decentralized.

\begin{itemize}
\item \textbf{Agents:} Heterogenous Agents with different speeds and different collision radius that are assigned a different sequence of target locations

\item \textbf{Targets:} Target locations with a circular keep-out radius that the agents are assigned to. The model takes in the sequence of target locations for each agent. This approach allows for the integration of a higher-level centralized controller that handles the Task Allocation and Scheduling\cite{messing_grstaps_2022} while allowing the Motion Planner to be run decentralized.

\item \textbf{Obstacles:} Obstacles or opponents that the agents need to avoid with a circular keep-out radius. These opponents can be moving or static. At each time step, the model takes in their current location and generates a new action for the agents.

\begin{figure}[tb]
\centering
\resizebox{\columnwidth}{!}{
\includegraphics{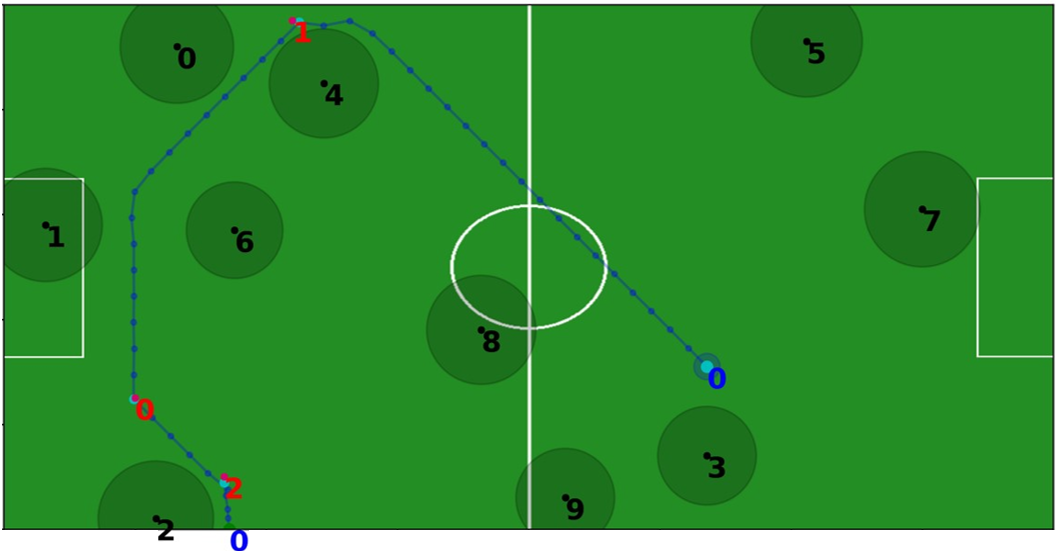}
}
\caption{Sample policy of the Trained Model blue agent avoiding static black Obstacles while moving from red Targets 1 to 0 to 2 in Robosoccer Domain with 0 collisions. }
\label{fig:sample_plan}
\vspace{-0.2in}
\end{figure}
\end{itemize}

\subsection{Problem Formulation}
We formulate the motion planning problem as a fully observable Markov Decision Process (MDP) using the five-tuple $\langle S, A, T, R, \gamma \rangle$ shown below:
\begin{itemize}
    \item States: The problem state $S$ in Motion Planning Problems is a joint state consisting of all the agent, target, and obstacle information.
    \item Actions: Actions at time-step $t$ within the Motion Planning Domain refers to a complete set of single-step motion plans for each agent, denoted as $A_t = [\langle \Delta x_{i_t}, \Delta y_{i_t} \rangle, \langle \Delta x_{i+1_t}, \Delta y_{i+1_t} \rangle, ...]$, to be executed at the same time in time-step $t$ for all of agents $i$.
    \item Transitions: T corresponds to executing the action in Motion Planner and proceeds to the next time step.
    \item Rewards: $R_t$ is based on the scheduling objective the user wants to optimize and the number of collisions that the agents have collided. In Section \ref{sec:Policy_Optimization} we show how to compute $R_t$ when optimizing motion plan to avoid collision with obstacles.
    \item Discount factor: $\gamma$.
\end{itemize}

\subsection{Model}
We utilize a simple Neural Network Model that takes in the following information to generate an output of $\langle \delta x, \delta y \rangle$.
\begin{itemize}
    \item \textbf{Agent Information:} Current location $\langle x_{u,t}, y_{u,t} \rangle$ at time-step $t$ for agent $u$.
    \item \textbf{Target Information:} Current location,  $\langle x_{v,t}, y_{v,t} \rangle$ at time-step $t$ for target $v$.
    \item \textbf{Obstacle Information:} A set of the current location, the keep-out radius, and distance from the agent's current location to its current location for each obstacle,  $[\langle x_{o_i,t}, y_{o_i,t}, r_{o_i}, d_{o_i,t} \rangle, \langle x_{o_{i+1},t}, y_{o_{i+1},t}, r_{o_{i+1}}, d_{o_i,t} \rangle, ...]$ at time-step $t$ for all $N$ closest obstacles $o_i$.
\end{itemize}

\paragraph{Baseline} We compare the performance of our results to a Target-to-Target Heuristic that moves to the location of the given target in a straight line without knowledge of the obstacle locations. This baseline is also used in the Dynamic Environment for the Obstacles, as they move to the current observed location of the Agents. It serves as an upper bound for our model in the number of target location it can reach.

\subsection{Policy Optimization} \label{sec:Policy_Optimization}
Our reward policy is trained using a Reward based on the distance to target, and the distance to the obstacles \cite{hausknecht_deep_2016} and is formulated in Equation \ref{eq:Reward_Function}.
\begin{align}
    R = - \alpha D(u,v) + \sum_{i=1}^N g(D(u,o_{i}),r_{o_i}) \label{eq:Reward_Function} \\
    g(D,r) = \begin{cases}
                \beta_1 (D-r),  \text{if } D-r \geq 0 \\
                \beta_2 (D-r)  \text{otherwise }
        \end{cases}
\end{align}
where $\alpha$, $\beta_1$, and $\beta_2$ are constants set to 10, 1, 100, respectively, and $D$ is the distance function to the obstacle's center and $r$ is the radius of the keep out range from the center where the unit is 'threatened' and collides with the obstacle.

We train our model in using Policy Gradient methods that seek to directly optimize the model parameters based on the present reward received from the environment \cite{peters_policy_2010}. 
\begin{equation}
    \label{eq:Loss Function}
    \begin{split}
    \nabla_\theta J(\theta) = \mathbb{E}_\pi ( \sum_{t}^{T} A_t^{\pi_\theta}(s_t, a_t) \nabla_\theta log\pi_\theta(a_i | s_t))
    \end{split}
\end{equation}

In Equation \ref{eq:Loss Function}, the advantage term, $A_t$, is estimated by subtracting the mean reward from a batch of policies from the reward of each policy. Each decision is sampled from a policy generated from a single observation, and an action is chosen randomly across them to perpetuate the environment in the next time step. We use the gradients calculated from Equation \ref{eq:Loss Function} to update the model weights.

\section{Experimental Results}
We train our model in a static environment for a single agent, 3 targets, and 10 obstacles with knowledge of $N=10$ closest obstacles for 4000 training steps. 
Our observation for the model is limited to observing the closest 10 obstacles. This is the limitation of the model being unable to scale as the number of obstacles increases.

We set $\gamma$ = 0.99, batch size = 8 and used Adam optimizer \cite{kingma_adam_2017} with a learning rate of $8 \times 10^{-3}$, and a weight decay of $1 \times 10^{-4}$. Specifically, we compute the gradient of the model using the log-likelihood at each stage for each agent, as shown in Equation \ref{eq:Loss Function}:
We train our model using static obstacles for 1 agent and 3 targets. All models are trained and evaluated on a Mac Studio 2022 with an Apple M1 Ultra Chip.

We test our model on different seeds (10, 11, 12) and compare against the Baseline on 3 domains for 3 independent sets of 100 test problems as follows:
\begin{itemize}
    \item \textbf{Static Simple Domain:} 1 Agent assigned to 3 Targets in random order with static 10 Obstacles.
    \item \textbf{Multi-Agent Dynamic Domain:} 3 Agents assigned to 3 Targets in random order, with mobile 10 Obstacles moving to one of the 3 Agents at random in a straight line, at the same speed as Agents.
    \item \textbf{Multi-Agent Dynamic Domain (Large):} 3 Agents assigned to 10 Targets in random order, with 10 mobile Obstacles that are moving to one of the 3 Agents at random in a straight line, at the same speed as Agents.
\end{itemize}

\begin{figure*}
\centering
\subfigure[]
{
\includegraphics[width=.275\linewidth]{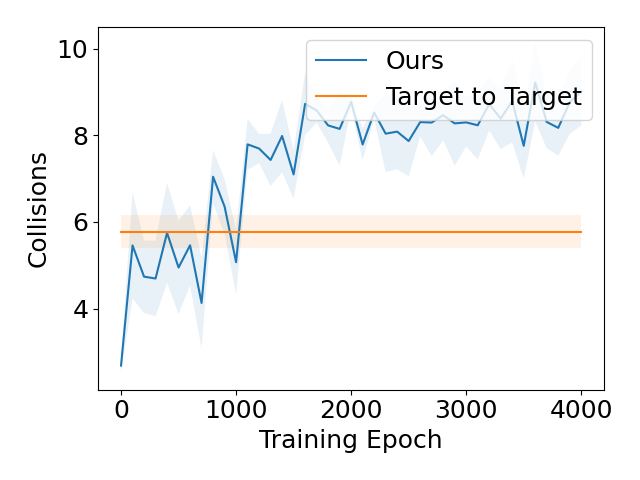}
\label{fig:simple_collision}
}
\subfigure[]{
\includegraphics[width=.275\linewidth]{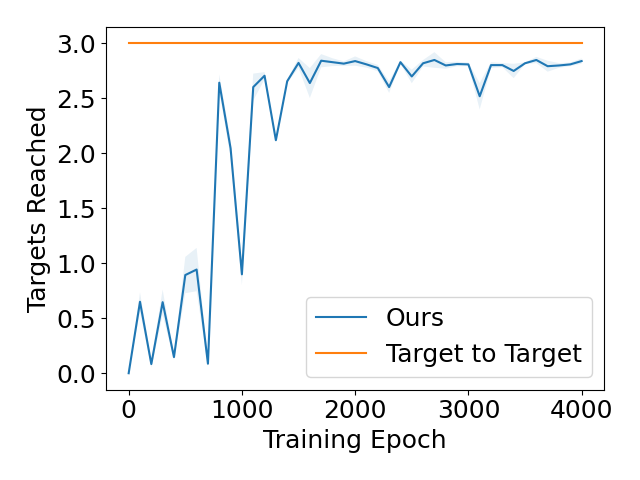}
\label{fig:simple_target}
}
\subfigure[]{
\includegraphics[width=.275\linewidth]{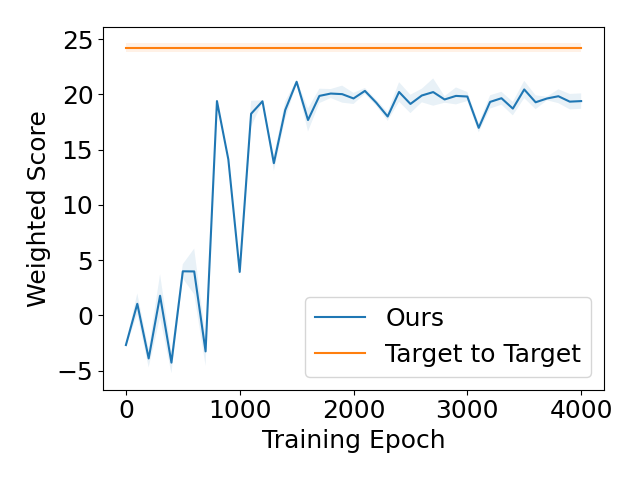}
\label{fig:simple_score}
}
\caption{Single-Agent Static Domain with 3 Targets, and 10 Obstacles Performance for Collisions (lower is better), Number of Targets reached, and Weighted Score based on Eq. \ref{eq:weighted_score} (higher is better).}
\label{fig:simple}
\end{figure*}

\begin{figure*}[]
\centering
\subfigure[]{
\includegraphics[width=.275\linewidth]{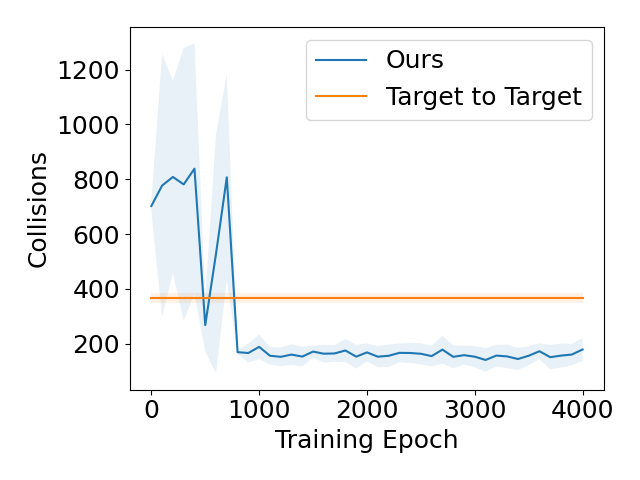}
\label{fig:moving_collision}
}
\subfigure[]{
\includegraphics[width=.275\linewidth]{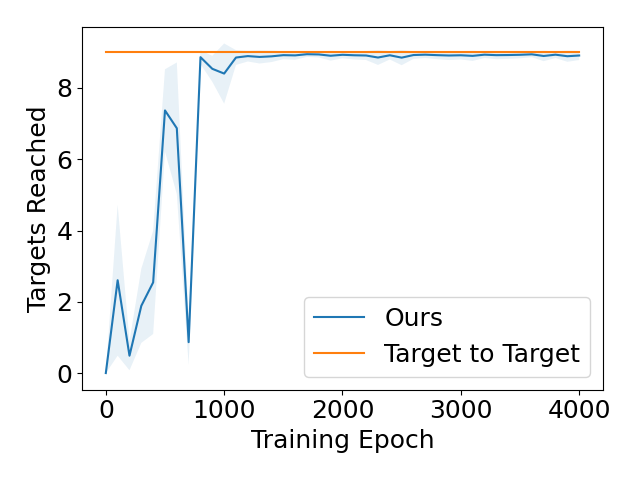}
\label{fig:moving_target}
}
\subfigure[]{
\includegraphics[width=.275\linewidth]{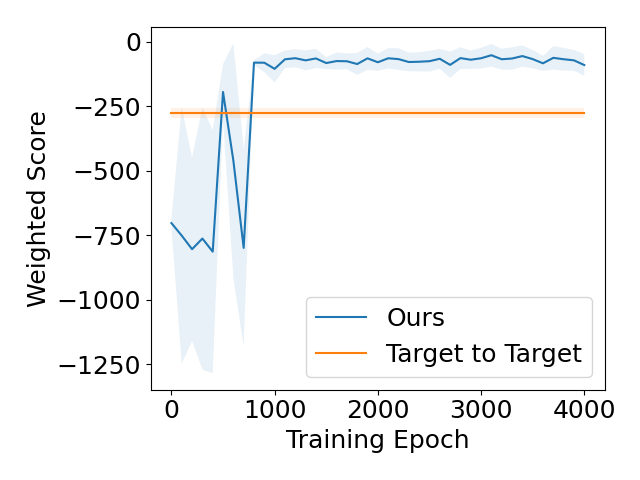}
\label{fig:moving_score}
}
\caption{Multi-Agent Dynamic Domain with 3 Agents, 3 Targets, and 10 Obstacles Performance for Collisions (lower is better), Number of Targets reached, and Weighted Score based on Eq. \ref{eq:weighted_score} (higher is better).}
\label{fig:moving}
\end{figure*}

\begin{figure*}[]
\centering
\subfigure[]{
\includegraphics[width=.275\linewidth]{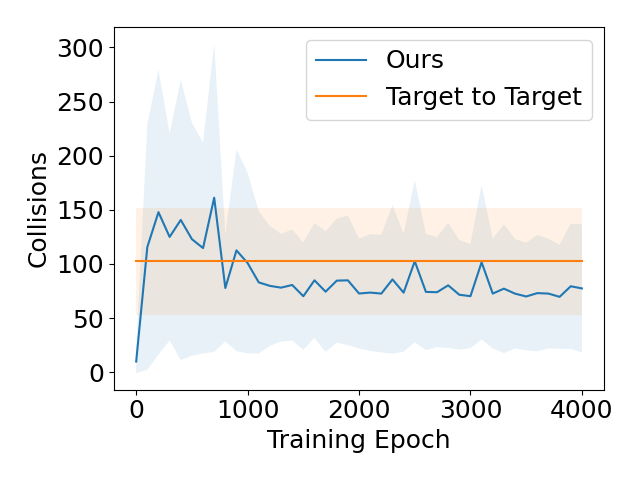}
\label{fig:partial_collision}
}
\subfigure[]{
\includegraphics[width=.275\linewidth]{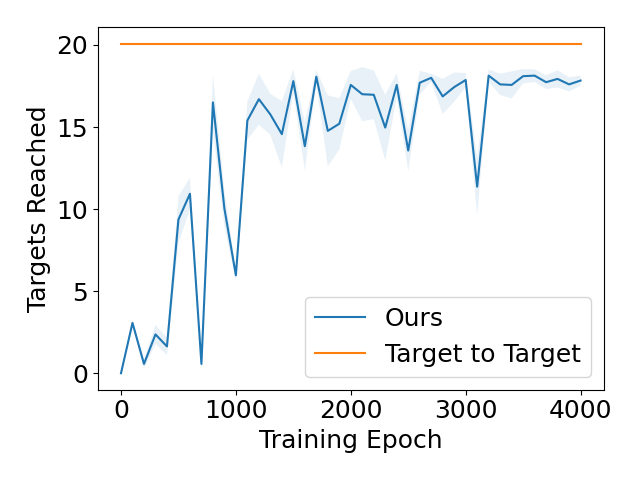}
\label{fig:partial_target}
}
\subfigure[]{
\includegraphics[width=.275\linewidth]{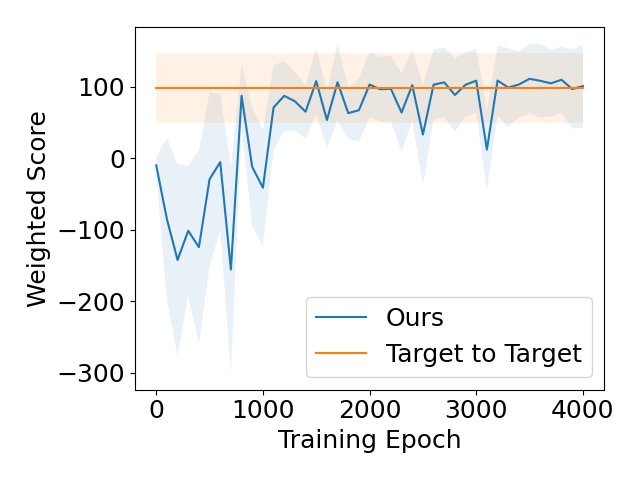}
\label{fig:partial_score}
}
\caption{Multi-Agent Dynamic Domain (Large) with 3 Agents, 10 Targets, and 10 Obstacles Performance for Collisions (lower is better), Number of Targets reached, and Weighted Score based on Eq. \ref{eq:weighted_score} (higher is better).}
\label{fig:partial}
\vspace{-0.2in}
\end{figure*}

We evaluate the performance of a Motion Planning Policy using 3 metrics:
\begin{itemize}
    \item Number of Collisions over the entire timeline of each problem instance, with the maximum performance based on the number of time steps.
    \item Number of Targets Reached over the entire runtime of the problem.
    \item Weighted Score based on: 
    \begin{equation}
        S = \alpha \times n_\tau - n_c
        \label{eq:weighted_score}
    \end{equation} where $\alpha$ is a constant set to $10$,  $n_\tau$ is the number of targets reached, and $n_c$ is the number of collisions over the entire runtime instance.
\end{itemize}
We show the performance of the trained obstacle avoidance policy compared to the baseline solution of going directly to the target, without any knowledge of the obstacles in Figures \ref{fig:simple}, \ref{fig:moving} and \ref{fig:partial}.


The baseline is always able to reach the provided targets due to taking the shortest path to the location in the continuous domain. The performance of our model in Figs. \ref{fig:simple_target}, \ref{fig:moving_target} and \ref{fig:partial_target} show that our model learns to reach the target locations.

In the \textbf{Simple Domain} in Figs. \ref{fig:simple}, the performance of the trained model is less than the performance of the baseline, as there are more collisions and the weighted score shows a similar pattern to the number of targets reached. The initial low number of collisions is due to the untrained policy, where the agent is not moving to the target and therefore not moving within the keep-out area of the obstacles.

In the \textbf{Multi-Agent Dynamic Domain} in Figs. \ref{fig:moving}, the number of collisions decreases over time as the mobile obstacles move to the location of the agents. This forces the agents to learn to move away. The learned performance shows that training learned on simple path planning in static environments is transferable to dynamic domains.

The main challenges of \textbf{Multi-Agent Dynamic Domain (Large)} are the number of Targets being too large for even the fastest policy (baseline) to complete, and the lack of full-observability for our model. The trained model takes in the 10 closest obstacles, treating Targets that it is not currently assigned to as obstacles as well. The limited observability of a larger number of obstacles leads to a decrease in the performance of the trained model as seen in Figs. \ref{fig:partial}.

\section{Conclusion and Future Work}
We present a Multi-Agent Agile Motion Planning Model for navigation across an environment with static and moving obstacles. We show that our model is end-to-end trainable in a custom environment that is based on Robosoccer, with multiple agents required to avoid moving obstacles. We further show that our model is capable of handling navigation in continuous space dynamically and avoiding moving targets.

A key limitations of the current model is its scalability. Scalability in multi-agent teams is a challenge that is an open research problem \cite{natarajan_human-robot_2023}. We plan to integrate the use of Graph Neural Networks for the Motion Planner to address this need for scalability. While our model scales with the number of agents, it is unable to accurately represent the observation space leading to performance drop as seen in Figure \ref{fig:partial}. The scalability of graph-based models would allow us to address the limitations presented within our results \cite{altundas_learning_2022, wang_heterogeneous_2022}.

The environment that we have presented utilizes a fully observable MDP. We show that our model is able to perform under partial observability or stochastic behavior of team members or opponents. We plan to expand on our current work to account for different observability conditions.

With the presence of an adversary, the environment we have presented can be further developed into accounting for Adversarial games, using the Game Theoretic approaches for MARL \cite{celli_coordination_2019}. We plan to further develop our environment to more accurately represent different adversarial gameplay scenarios.

\bibliographystyle{IEEEtran}
\bibliography{workshop}

\end{document}